\documentclass[runningheads]{llncs}
\usepackage[T1]{fontenc}
\usepackage{graphicx}
\usepackage{float}
\usepackage{multirow}
\usepackage{multicol}
\usepackage{url}
\begin{document}
\title{Impact of Synthetic Images on Morphing Attack Detection Using a Siamese Network}
\titlerunning{Impact of Synthetic Images on MAD}
%
\author{Juan Tapia \and Christoph Bush \inst{1}}
\authorrunning{Tapia et al.}
%
\institute{da/sec-Biometrics and Internet Security Research Group, Hochschule Darmstadt, Germany. 
\email{juan.tapia-farias, christoph.busch{@h-da.de} }\\
}
\maketitle              
\begin{abstract}
This paper evaluated the impact of synthetic images on Morphing Attack Detection (MAD) using a Siamese network with a semi-hard-loss function. Intra and cross-dataset evaluations were performed to measure synthetic image generalisation capabilities using a cross-dataset for evaluation. Three different pre-trained networks were used as feature extractors from traditional MobileNetV2, MobileNetV3 and EfficientNetB0. Our results show that MAD trained on EfficientNetB0 from FERET, FRGCv2, and FRLL can reach a lower error rate in comparison with SOTA. Conversely, worse performances were reached when the system was trained only with synthetic images. A mixed approach (synthetic + digital) database may help to improve MAD and reduce the error rate. This fact shows that we still need to keep going with our efforts to include synthetic images in the training process.

\end{abstract}

\section{Introduction}
Face morphing can be understood as an algorithm that combines two or more look-alike facial images from one subject and an accomplice who could apply for a valid passport exploiting the accomplice's identity. Morphing takes place in the enrolment process stage. The threat of morphing attacks is known for border crossing or identification control scenarios. Efforts to spot such attacks can be broadly divided into two types: (1) Single Image Morphing Attack Detection (S-MAD) techniques and Differential Morphing Attack Detection (D-MAD) methods. 

S-MAD can detect both landmark-based morphing and synthetic GAN-based morphing methods \cite{Morgan}, \cite{MIPGAN} while integrating a variety of different feature extraction methods observing texture, shape, quality, residual noise, and others \cite{TapiaSMAD}. 

In an operational scenario, a manual passport or ID card inspection complements this automatic classification at border control gates to detect morphing attacks when a suspicious image is presented or the border police have received previous advice. Today, MAD is still an open challenge to develop generalisation capabilities. The following open issues can be determined according to the state-of-the-art:  Cross-Dataset (CD), Cross-Morphing (CM), Leave-One-Out (LOO) evaluations, and a reduced number of bona fide images.

In order to train a robust MAD method (single or differential) is complicated due to the lack of a large-scale database available for this purpose. Only some existing databases allow us to create Morphing Attacks. It is essential to highlight that the image only can be used for the purpose informed in the consent form. Currently, most of the papers in the literature use FERET and FRGCv2 datasets to generate images with several morphing tools. Other open-access databases, such as FRLL and the London dataset, can be used directly with many Morphing Attacks. However, these databases have available only 204 subjects as bona fide images. 102 images are neutral expressions and the same subject (102 subjects) with a smiling expression. Overall, we have an unbalanced dataset. 

On the other hand, capturing images from real Automatic Border Control (ABC) gates is a difficult task because of the equipment and the authorisation to perform this task in real scenarios. It is not trivial to capture sessions for many researchers. 

This paper aims to complement previous approaches and analyse the impact of using purely synthetic images in a cross-dataset scenario. This means training with bona fide and morph face images (both from synthetic images) and evaluating in a benchmark test-set and assessment also in the state-of-the-art databases based on morph created from digital images and viceversa. 
Those databases contain morphed images created using landmark and GANs-based morphing images. This cross-evaluation also considers different morphing tools and may help focus on the next steps of MAD. In summary, the main contributions of this paper are:

\begin{itemize}
    \item This MA analysis complements the current state-of-the-art. An intra-dataset analysis was performed, training in synthetic face images (bona fide/morph) and testing purely on synthetic images.
    \item  A cross-dataset evaluation was performed using purely synthetic images for training and evaluation on State-Of-The-Art (SOTA) digital databases such as FERET, FRGC, FRLL and others. 
    \item  A cross-dataset evaluation was performed using a mix-dataset (digital + synthetic) for training and evaluation on State-Of-The-Art (SOTA) digital databases such as FERET, FRGC, FRLL and others. 
    \item Several Morphing tools were used in order to explore different qualities of morph images using a Siamese network based on a semi-hard triplet loss to improve the MAD accuracy.
    \item  Our results outperform the SOTA on synthetic images but show that using only synthetic images as bona fice and morph is still not enough to detect real MAD based on digital images. Conversely, a system trained in SOTA or mixed can detect synthetic images with a low error rate.

\end{itemize}

The rest of the manuscript is organised as follows: Section~\ref{sec:relate} summarises the related works on MAD. The database description is explained in Section\ref{sec:database}. The metrics are explained in Section~\ref{sec:metric}. The experiment and results framework is then presented in Section~\ref{sec:exp_results}. We conclude the article in Section~\ref{sec:conclusions}.

\section{Related Work}
\label{sec:relate}

Due to legal and privacy issues, the use of face image data collected from the web is problematic for research purposes. Privacy regulations such as the GDPR assure individuals the right to withdraw their consent to use or store their private data, practically making the use of large face datasets difficult. This restriction and the difficulty of capturing images in a real-time process from  ABC gates encourage the research community to develop new morphing images (bona fide and morphed) from Generative Adversarial Networks (GAN) to create synthetic face databases \cite{Naser-privacy}. 

Damer et al. \cite{Naser-privacy} raised the question: "can morphing attack detection (MAD) solutions be successfully developed based on synthetic data?". Towards that, it introduced the first synthetic-based MAD development dataset, the Synthetic Morphing Attack Detection Development dataset (SMDD). 

Huber et al. \cite{SDD} conducted a Competition on Face Morphing Attack Detection Based on Privacy-aware Synthetic Training Data. The competition was held at the International Joint Conference on Biometrics 2022. A new benchmark dataset was released called SYN-MAD-2022. The benchmark is based on the Face Research Lab London dataset (FRLL). In this competition, the best results according to ranking were reached by the MorphHRNet team obtained very good results. However, this team used a mixture of synthetic images with traditional FRLL, FERET and FRGC databases for evaluation. Then, the real impact of using only synthetic images was not explored and tested for all the teams.

Related to the Siamese network on MAD, previous work has been proposed based on deep learning. Borgui et al.~\cite{siamese-guido} proposed a differential morph attack detection based on a double Siamese architecture.
The proposed framework consisted of two different modules, referred to as “Identity” and 'Artefact blocks', respectively, and each block was based on a Siamese network followed by a Multi-Layer Perceptron (MLP) that acts as fusion layers. Finally, a Fully Connected layer (FC) merges the features originating from the two modules and outputs the final score. Experimental results were obtained in three datasets: PMDB, MorphDB, and AMSL. This approach used a Contrastive loss.

Soleymani et al.~\cite{Soleymani} developed a novel differential morphing attack detection algorithm using a deep siamese network. The Siamese network takes image pairs as inputs and yields a score on the likelihood that the face images are from the same subject. They employ a pre-trained Inception ResNetv1 as the base network initialised with weights pre-trained on VGGFace2. The experiments are conducted on two separate morphed image datasets: VISAPP17 and MorGAN. This approach used a contrastive loss.

Chaudhary et al.~\cite{Chaudhary_2021_CVPR} proposed a differential morph attack detection algorithm using an undecimated 2D Discrete Wavelet Transform (DWT). By decomposing an image to wavelet sub-bands, we can more clearly identify the morph artefacts hidden in the image domain in the spatial frequency domain.

\section{Database}
\label{sec:database}
In order to measure the real impact of synthetic images, several morphing databases have been used. The description of each one is detailed as follows:

\subsection{SYN-MAD-2022}
This dataset was developed for the Competition on Face Morphing Attack Detection Based on Privacy-aware Synthetic Training Data (SYN-MAD) \cite{SDD}. It is divided into a training set of 25k bona fide and 15k morphed images based on synthetic faces, which avoids using privacy-sensitive real-face images. Furthermore, the images bounding box and five facial landmark points are provided.
The test set, named MAD evaluation benchmark database (MAD22), was created by the organisers as part of the competition, and it is publicly available \footnote{\url{https://github.com/marcohuber/SYN-MAD-2022}}. 

The benchmark test set (SDD) contains 4,483 divided into 984 OpenCV, 1,000 FaceMorpher, 500 Webmorph, 1,000 MIPGAN-I, 999 MIPGAN-II morphed face images and 204 bona fide images from the FRLL dataset. 
It is essential to highlight that even using similar morphing tools in some cases, such as FaceMorpher, OpenCV and WebMorph, the resulting images are not the same as the original FRLL databases because the subjects were combined in a different order. 

\subsection{SOTA Databases}
In this paper, four different databases of frontal faces images were also used: the Facial Recognition Technology (FERET)~\cite{Phillips-FERET-1998}, the Face Recognition Grand Challenge (FRGCv2)~\cite{PhillipsFRGC}, the  Face Research London Lab (FRLL)\cite{FRLL} and AMSL database \cite{debruine2017face}. The morphed images in these datasets have been created using a morphing factor of 0.5, meaning both subject images contribute equally to the morphed image. 
The FRLL and AMSL morphed images have been generated with the following morphing tools: FaceMorpher, FaceFusion and WebMorpher based on landmarks and StyleGAN from FRLL without landmarks. The description of each database is explained as follows:

The FERET dataset is a subset of the Colour FERET Database, generated in the context of the Facial Recognition Technology program technically handled by the National Institute of Standards and Technology (NIST). It contains 569 bona fide face images.

The FRGCv2 dataset used in this work is a constrained subset of the second version of the Face Recognition Grand Challenge dataset. It contains 979 bona fide face images.

The FRLL dataset is a subset of the publicly available Face Research London Lab dataset. It contains 102 bona fide neutral and 102 smiling images. Three morphing algorithms were applied to obtain 1,222 morphs from the FaceMorpher algorithm, 1,222 morphs from the StyleGAN algorithm, and 1,222 morphs from the WebMorph algorithm \cite{webmorph}.

\begin{figure}
\centering
\includegraphics[scale=0.50]{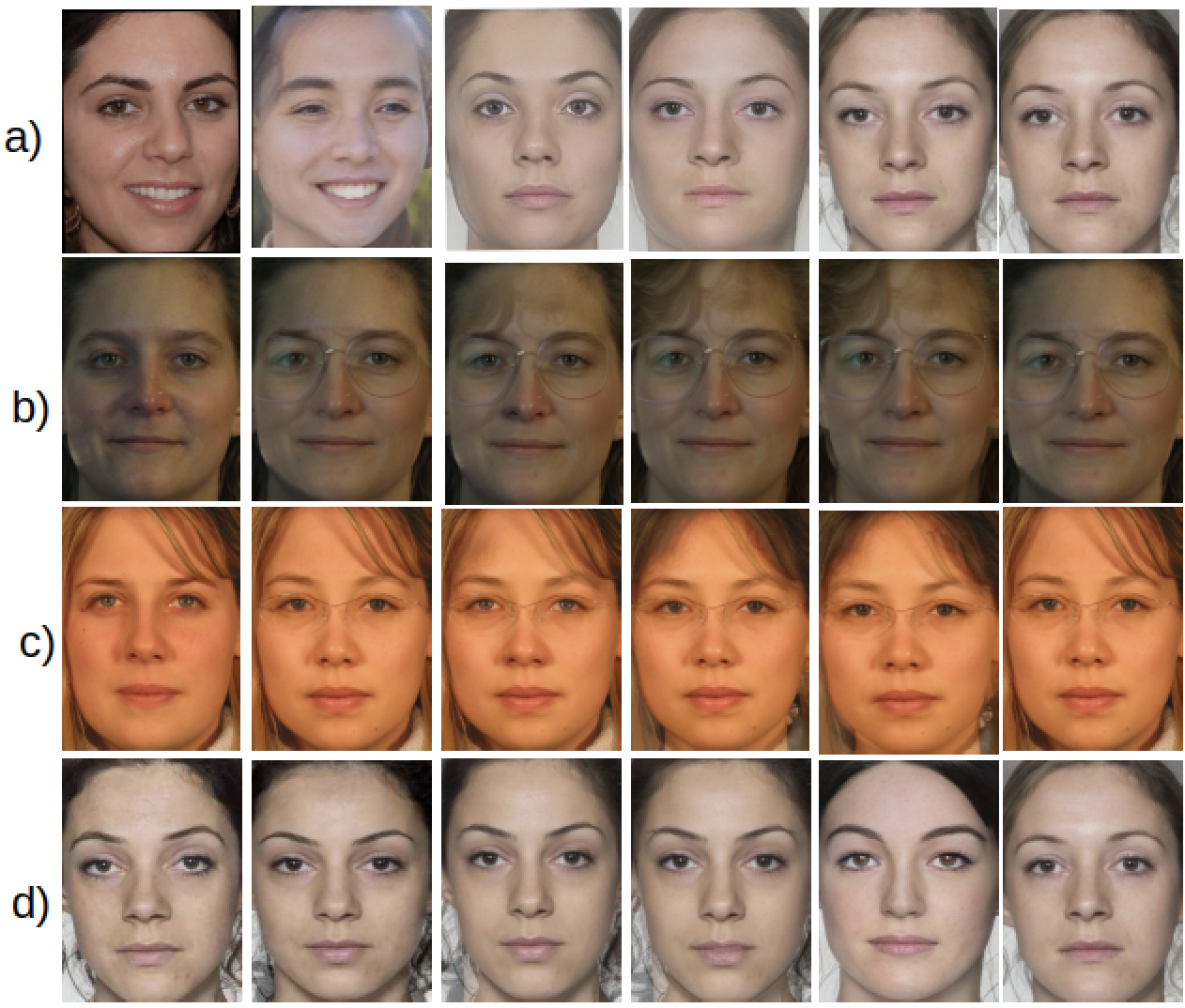}
\caption{\label{sum_images} Example of images from all the databases. a) SDD: Left to Right: Bona fide (Subject-1) synthetic, Morph Synthetic (Subject-2), MIPGAN-I, MIPGAN-II, OpenCV-Morpher, WebMorpher. B) FERET: Left to right: Bona fide (Subject-1), Bona fide (Subject-2), FaceMorpher, FaceFusion, OpenCV-Morpher, UBO-Morpher. c) FRGCv2: Left to right: Bona fide (Subject-1), Bona fide (Subject-2), FaceMorpher, FaceFusion, OpenCV-Morpher, UBO-Morpher. d) FRLL: Left to Right: Bona fide (Subject-1), Bona fide (Subject-2), AMSL, FaceMorpher, OpenCV-Morpher, and StyleGAN.}
\vspace{-1.5em}
\end{figure}

The AMSL Face Morph Image Data Set is a collection of bona fide and morphed face images that can be used to evaluate the detection performance of MAD algorithms. The images are organised as follows: genuine-neutral with 102 genuine neutral face images, genuine-smiling with 102 genuine smiling face images and 2,175 morphing face images. Figure \ref{sum_images} shows a side-by-side image example of all the databases used in this work \footnote{All the image sources have been properly cited throughout the paper}.
Table \ref{tab:summary_db} shows a summary of all the databases used in this work.

\begin{table}[]
\centering
\scriptsize
\caption{Summary databases}
\label{tab:summary_db}
\begin{tabular}{|c|c|c|c|c|c|}
\hline
Database & Bona fide & Morph  & Type       & Tool       & Notes \\ \hline
AMSL     & 204      &  2,175 & Landmark   & WebMorph   &  The same bona fide images as FRLL.     \\ \hline
\multirow{4}{*}{FRLL} & 
  \multirow{4}{*}{204} &
  1,222 &
  Landmark &
  OpenCV &
  \multirow{4}{*}{} \\ \cline{3-5}
         &          & 1,222 & Landmark   & FaceMorpher &       \\ \cline{3-5}
         &          & 1,222 & Synthetic & StyleGAN2   &       \\ \cline{3-5}
         &          & 1,222 & Landmark   & WebMorpher  & The same bona fide images as AMSL.      \\ \hline
\multirow{4}{*}{FERET} &
  \multirow{4}{*}{529} &
  529*3 &
  Landmark &
  FaceFusion &
  \multirow{4}{*}{\begin{tabular}[c]{@{}c@{}}(*) Print/Scan 300dpi,\\ Print/Scan 600dpi,\\ Digital-Resize (No P/S)\end{tabular}} \\ \cline{3-5}
         &          & 529*3 & Landmark   & FaceMorpher &       \\ \cline{3-5}
         &          & 529*3 & Landmark   & OpenCV      &       \\ \cline{3-5}
         &          & 529*3 & Landmark   & UBO-Morpher &       \\ \hline
\multirow{4}{*}{FRGC} &
  \multirow{4}{*}{979} &
  979*3 &
  Landmark &
  FaceFusion &
  \multirow{4}{*}{\begin{tabular}[c]{@{}c@{}}(*)Print/Scan 300dpi,\\ Print/Scan 600dpi,\\ Digital-Resize (No P/S)\end{tabular}} \\ \cline{3-5}
         &          & 979*3 & Landmark   & FaceMorpher &       \\ \cline{3-5}
         &          & 979*3 & Landmark   & OpenCV      &       \\ \cline{3-5}
         &          & 979*3 & Landmark   & UBO-Morpher &       \\ \hline
  SYD-MAD &
  15.000 + 4,948* & 25k  &  Synthetic & N/A &
  \begin{tabular}[c]{@{}c@{}}(*)Benchmark Test set contains:\\ OpenCV, FaceMorpher, WebMorpher,\\ MIPGAN-I, MIPGAN II.\end{tabular} \\ \hline
\end{tabular}
\vspace{-1.5em}
\end{table}

According to the SOTA, different algorithms were used to create morph images. The following six morphing tools have been used:

\begin{itemize}
    \item \textbf{FaceFusion}~\cite{facefusion}: this proprietary mobile application developed by MOMENT generates realistic faces since morphing artefacts are almost invisible.
    \item \textbf{FaceMorpher}~\cite{facemorpher}: this open-source Python implementation relies on STASM, a facial feature finding the package, for landmark detection, but generated morphs show many artefacts which make them more recognisable.
    \item \textbf{OpenCV-Morpher}~\cite{opencv_library}: this open-source morphing algorithm is similar to the FaceMorpher method but uses Dlib to detect face landmarks. Again, some artefacts remain in generated morphs.
    \item \textbf{UBO-Morpher}~\cite{UBO}: The University of Bologna developed this algorithm. The resulting images are of high quality without artefacts in the background.
    \item \textbf{WebMorpher}~\cite{webmorph}: this open-source morphing algorithm is a web-based version of Psychomorph with several additional functions. While WebMorph is optimised for averaging and transforming faces, you can delineate and average any image.
    \item \textbf{StyleGAN2}~\cite{stylegan}: this open-source morphing algorithm by NVIDIA, No landmarks are used to create morph images.
\end{itemize}

\subsection{Metrics}
\label{sec:metric}

The detection performance of the investigated S-MAD algorithms was measured according to ISO/IEC 30107-3 \footnote{\url{hhttps://www.iso.org/standard/79520.html}} using the Equal Error Rate (EER), Bona fide Presentation Classification Error Rate (BPCER), and Attack Presentation Classification Error Rate (APCER) metric defined as (\ref{eq:bpcer}) and (\ref{eq:apcer}).

\begin{equation}\label{eq:bpcer}
    BPCER=\frac{\sum_{i=1}^{N_{BF}}RES_{i}}{N_{BF}}
\end{equation}

\begin{equation}\label{eq:apcer}
    APCER=\frac{1}{N_{PAIS}}\sum_{i=1}^{N_{PAIS}}(1-RES_{i})
\end{equation}

Where $N_{BF}$ is the number of bona fide presentations, $N_{PAIS}$ is the number of morphing attacks for a given attack instrument species and $RES_{i}$ is $1$ if the system's response to the $i-th$ attack is classified as an attack and $0$ if classified as bona fide. In this work, S-MAD performance is reported using the Equal Error Rate (EER), the point which corresponds to relevant security settings where the APCER is equal to BPCER. Also, two operational points are reported BPCER10 and BPCER20. The BPCER20 is the BPCER value obtained when the APCER is fixed at 5\%, and BPCER10 (APCER at 10\%).

\section{Method}
A Siamese network consists of two identical networks which can process different inputs. They are joined at their output layers based on the pre-trained networks by a unique function that calculates a metric between the embedding estimated by each network. Most of these networks used have been trained on ImageNet 1k weights. Because of that, we explored using 21k weights and also reduced the 21k to 1k weights. This kind of network is ideal for morphing attack detection because they are primarily designed to find similarities between two inputs. As a part of the process, a pre-computed template of four random bona fide images is processed in order to compare the embedding distances of the new input face image, which could be potentially morphed. 

A contrastive loss is typically used to train a siamese network to distinguish between mated (bona fide) and non-mated (morphed) pairs. Traditionally, the contrastive loss function optimises two identical Convolutional Neural Networks outputs, each operating on a different input image and using an Euclidean distance measure or an SVM classifier to make the final decision. At the same time, contrastive representations have achieved state-of-the-art performance on visual recognition tasks and have been theoretically proven effective for binary classification. However, according to \cite{FSL-tapia}, the triple-loss function could separate the morphed images more effectively based on the semi-hard triplet loss function to separate several morphing tools used for the same pairs of subjects. Then, the contrastive loss is not able to separate this kind of image because we have images from the same subject with several morphing tools. Conversely, triplet-loss can explore and deal very well with easy, semi-hard and hard examples.

\begin{figure}[H]
\centering 
\includegraphics[scale=0.30]{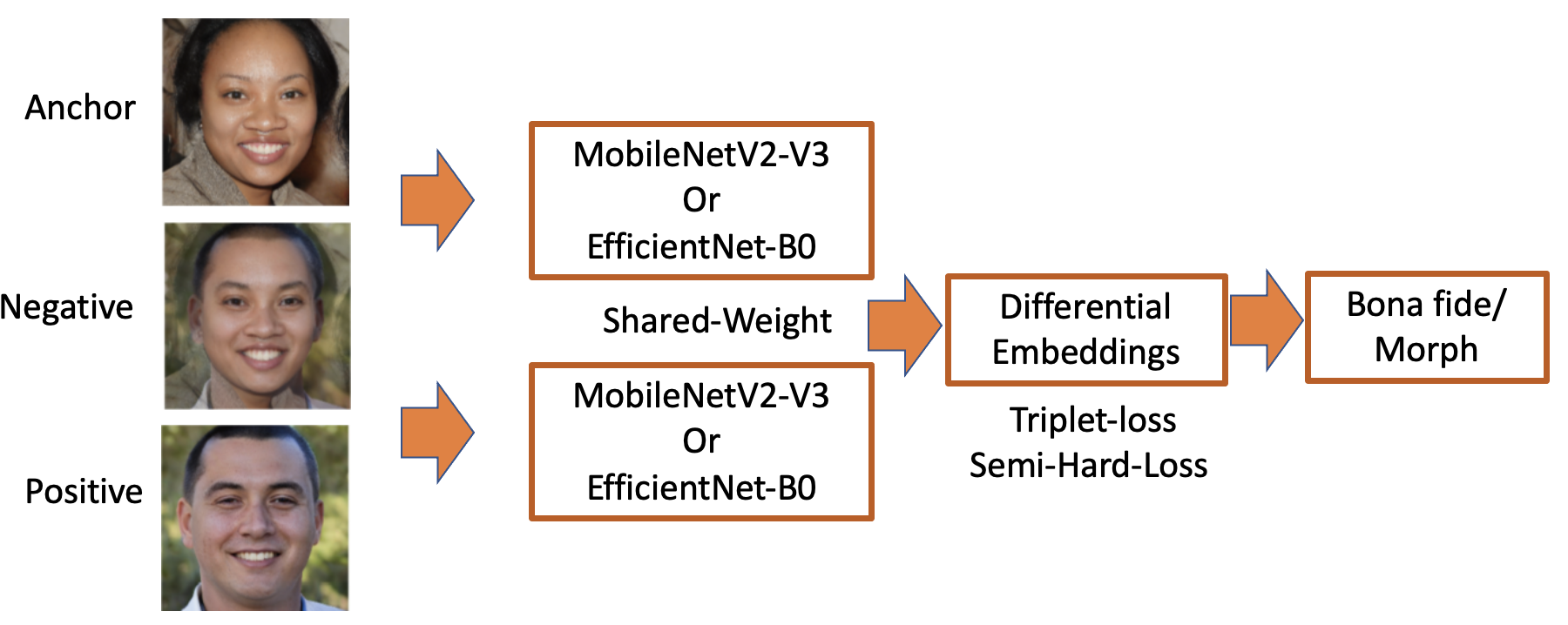}
\caption{\label{fig:siames} Siamese network.}
\vspace{-1.5em}
\end{figure}

In order to extract the embedding from the bona fide and morphed images, ImageNet (1k and 21k) pre-trained general-purpose backbones such as MobileNetV2\cite{SandlerMobileNet}, MobileNetV3\cite{MobilenetV3} and EfficientNetB0 \cite{EfficientNet} were used. Then, the network is retrained with a morphing database. Afterwards, the model is optimised by enforcing the triple loss function to measure the triplets relationships between both classes in a binary problem or $N$ classes in a multi-class problem. Figure \ref{fig:siames} shows a Siameses Network.


\subsubsection{Triplet Loss}
\label{TL}

A triplet loss function was used to train a neural network to closely embed features of the same class while maximising the distance between embeddings of different classes. An anchor (bona fide) and one negative (potentially morphed), and one positive sample (bona fide or same class of anchor) are chosen to do this~\cite{SchroffKP15}.

To formalise this requirement, a loss function is defined over triplets of embeddings:

\begin{itemize}
    \item An anchor image (a) - For example, bona fide.
    \item A positive image of (p) the same class as the anchor.
    \item A negative image (n) of a different class - In our example, a morphed image.
\end{itemize}

For some distance $(d)$ on the embedding space, the loss of a triplet $(a,p,n)$ is defined as:
\begin{equation}
    L_t = \max(d(a,p)-d(a,n)+margin,0)
\end{equation}

We minimise this loss, which pushes $d(a,p)$ to $0$ and $d(a,n)$ to be greater than $d(a,p)+margin$. As soon as $n$  becomes an “easy negative”, the loss becomes zero. We used a semi-hard triplet, which means that the negative is not closer to the anchor than the positive, but which still has a positive loss: 

\begin{equation}
d(a,p)<d(a,n)<d(a,p)+margin  
\end{equation}

\section{Experiments and Results}\label{sec:exp_results}
\vspace{-0.4em}

Four experiments are conducted based on the Siamese network. A semi-hard triple loss function applied to morphing attack detection using MobileNetV2, MobileNetV3 and EfficientNetB0 was explored. 

The morphing images were created based on OpenCV-Morphing tools for the SYN-MAD database. All the processes for pair selection and others were described in \cite{SDD}. 
For the images from AMSL, FRGC, FERET, and FRLL databases, the faces were detected and aligned based on MTCNN \cite{MTCNN}. For the SYN-MAD database, the coordinates of the faces are provided to crop the images. 

For all the experiments, the optimisation function used is Adam; even were explored SGD and RMSprop, Adam reached the best results. A grid search was used to identify the best learning rate (lr) for each experiment. Further on that, 250 epochs, batch size of 128 on a 21GB-NVidia-GPU were used.
An intra-dataset was performed in Experiment \#1 using the different configurations. For Experiments \#2 \#3 and \#4, a cross-dataset protocol was explored as explained as follows:

\subsection{Exp1 - Train SYN-MAD/TEST SDD-Bechmark}
For this experiment, three tests were conducted based on MobileNetV2 (ImagiNet-1k), MobileNetV3 (ImagiNet-21k), and EfficientNetB0 (ImagiNet-21k-1k). The process used the SYN-MAD release's official training and test set. 

Test a) Both MobileNetV2 networks were trained from scratch using the ImagiNet weight from 1,000 classes and adapted to classify bona fide and morphed images. 

Test b) Both MobileNetV3-small networks were trained from scratch using the ImagiNet weight from 21,000 classes and adapted to classify bona fide and morphed images.

Test c), Both EfficientNetB0 networks were trained from scratch using the ImagiNet weight from 21,000 classes but reduced to 1,000 as adapted to a binary problem to classify bona fide and morphed images. 

For the tests a), b) and c), 23,000 bona fide and 13,000 morphed synthetic images were used to train. For the validation set, 4,000 (2,000 bona fide and 2,000 synthetic) images were used.

The test c) with EfficientNetB0 reached the best performance. Table \ref{tab:results_sdd_our} shows a summary with the best results for experiment 1. For the SDD competition, we include the best results reported for each dataset. 

These results can be used as a reference for evaluation. Still, it is essential to highlight that some of the equipment uses mixtures of synthetic and digital images in the evaluation process. Our method, based only on synthetic and the Siamese network, reached competitive results compared with the best result on similar E-CBAM-VCMI and MorphHRNet team conditions. Table \ref{tab:results_sdd_our} shows the summary results.

\begin{table}[]
\centering
\caption{Summary evaluation report based on the SYD-MAD and our proposal.}
\label{tab:results_sdd_our}
\resizebox{\columnwidth}{!}{%
\begin{tabular}{|cccccccc|}
\hline
\multirow{2}{*}{Database}                                          & \multicolumn{3}{c}{IJCB 2022 -SMDD} & \multicolumn{3}{c}{Our Proposal}        & \multirow{2}{*}{Note} \\ \cline{2-7}
 &
  \begin{tabular}[c]{@{}c@{}}EER\\ (\%)\end{tabular} &
  \begin{tabular}[c]{@{}c@{}}BPCER10\\ (\%)\end{tabular} &
  \begin{tabular}[c]{@{}c@{}}BPCER20\\ (\%)\end{tabular} &
  \begin{tabular}[c]{@{}c@{}}EER\\ (\%)\end{tabular} &
  \begin{tabular}[c]{@{}c@{}}BPCER20\\ (\%)\end{tabular} &
  \begin{tabular}[c]{@{}c@{}}BPCER10\\ (\%)\end{tabular} &
   \\ \hline
\multicolumn{8}{|c|}{Experiment 1}                                                                                                                                         \\ \hline
OpenCV &
  27.54 &
  50.49 &
  38.24 &
  22.06 &
  33.59 &
  49.80 &
  \multirow{5}{*}{\begin{tabular}[c]{@{}c@{}}(90037)\\ \\ Trained:SDD Train-Set\\ Test: Benchmark Test-Set\\ Synthetic and Landmark\\ Compared with\\ E-CBAM@VCMI\end{tabular}} \\ \cline{1-7}
FaceMorpher                                                        & 41.20      & 95.10      & 82.35     & 2.00        & 0.00        & 0.00        &                       \\ \cline{1-7}
WebMorpher                                                         & 30.60      & 47.06      & 38.24     & 26.39       & 46.14       & 55.80       &                       \\ \cline{1-7}
MIPGAN-I                                                           & 32.50      & 53.43      & 40.69     & 11.76       & 13.23       & 22,07       &                       \\ \cline{1-7}
MIPGAN-II                                                          & 25.93      & 30.39      & 30.39     & 8.82        & 8.83        & 10.78       &                       \\ \hline
\multicolumn{8}{|c|}{Experiment 2}                                                                                                                                         \\ \hline
AMSL &
  \multicolumn{3}{c}{\multirow{5}{*}{N/A}} &
  3.70 &
  2.02 &
  3.29 &
  \multirow{5}{*}{\begin{tabular}[c]{@{}c@{}}(212642)\\ Trained:SDD Train-Set\\ Test: SOTA\end{tabular}} \\ \cline{1-1} \cline{5-7}
\begin{tabular}[c]{@{}c@{}}OpenCV\\ (FERET/FRGC)\end{tabular}      & \multicolumn{3}{c}{}                & 21.55/19.25 & 41.39/21.45 & 63.53/22.15 &                       \\ \cline{1-1} \cline{5-7}
\begin{tabular}[c]{@{}c@{}}FaceMorpher\\ (FERET/FRGC)\end{tabular} & \multicolumn{3}{c}{}                & 18.52/16.25 & 34.97/22.70 & 51.22/22.35 &                       \\ \cline{1-1} \cline{5-7}
\begin{tabular}[c]{@{}c@{}}FaceFusion\\ (FERET/FRGC)\end{tabular}  & \multicolumn{3}{c}{}                & 35.70/16.85 & 76.77/55.20 & 91.30/57.30 &                       \\ \cline{1-1} \cline{5-7}
\begin{tabular}[c]{@{}c@{}}UBO-Morpher\\ (FERET/FRGC)\end{tabular} & \multicolumn{3}{c}{}                & 32.70/19.65 & 70.69/55.50 & 81.47/61.25 &                       \\ \hline
\multicolumn{8}{|c|}{Experiment 3}                                                                                                                                         \\ \hline
OpenCV &
  \multicolumn{3}{c}{\multirow{5}{*}{N/A}} &
  16.66 &
  28.92 &
  54.51 &
  \multirow{5}{*}{\begin{tabular}[c]{@{}c@{}}(134122)\\ Trained: Train-SOTA\\ No synthetic images\\ Test: Benchmark Test-set\end{tabular}} \\ \cline{1-1} \cline{5-7}
FaceMorpher                                                        & \multicolumn{3}{c}{}                & 3.43        & 1.83        & 1.94        &                       \\ \cline{1-1} \cline{5-7}
WebMorpher                                                         & \multicolumn{3}{c}{}                & 33.4        & 74.50       & 80.71       &                       \\ \cline{1-1} \cline{5-7}
MIPGAN-I                                                           & \multicolumn{3}{c}{}                & 18.0        & 31.37       & 61.29       &                       \\ \cline{1-1} \cline{5-7}
MIPGAN-II                                                          & \multicolumn{3}{c}{}                & 8.33        & 6.86        & 16.66       &                       \\ \hline
\multicolumn{8}{|c|}{Experiment 4}                                                                                                                                         \\ \hline
OpenCV &
  5.69 &
  1.96 &
  1.47 &
  6.37 &
  1.90 &
  0.84 &
  \multirow{5}{*}{\begin{tabular}[c]{@{}c@{}}Trained: Mix \\ (synthetic plus SOTA)\\ Test: Benchmark Test-set\\ Compared with \\ MorphHRNet\end{tabular}} \\ \cline{1-7}
FaceMorpher                                                        & 5.90       & 1.96       & 1.47      & 3.50        & 0.48        & 0.98        &                       \\ \cline{1-7}
WebMorpher                                                         & 9.80       & 10.78      & 3.93      & 9.31        & 8.33        & 16.17       &                       \\ \cline{1-7}
MIPGAN-I                                                           & 15.30      & 24.22      & 11.27     & 4.90        & 1.96        & 4.41        &                       \\ \cline{1-7}
MIPGAN-II                                                          & 10.41      & 11.27      & 2.94      & 3.43        & 0.98        & 1.96        &                       \\ \hline
\end{tabular}%
}
\end{table}

\begin{figure}[H]
\centering 
\includegraphics[scale=0.24]{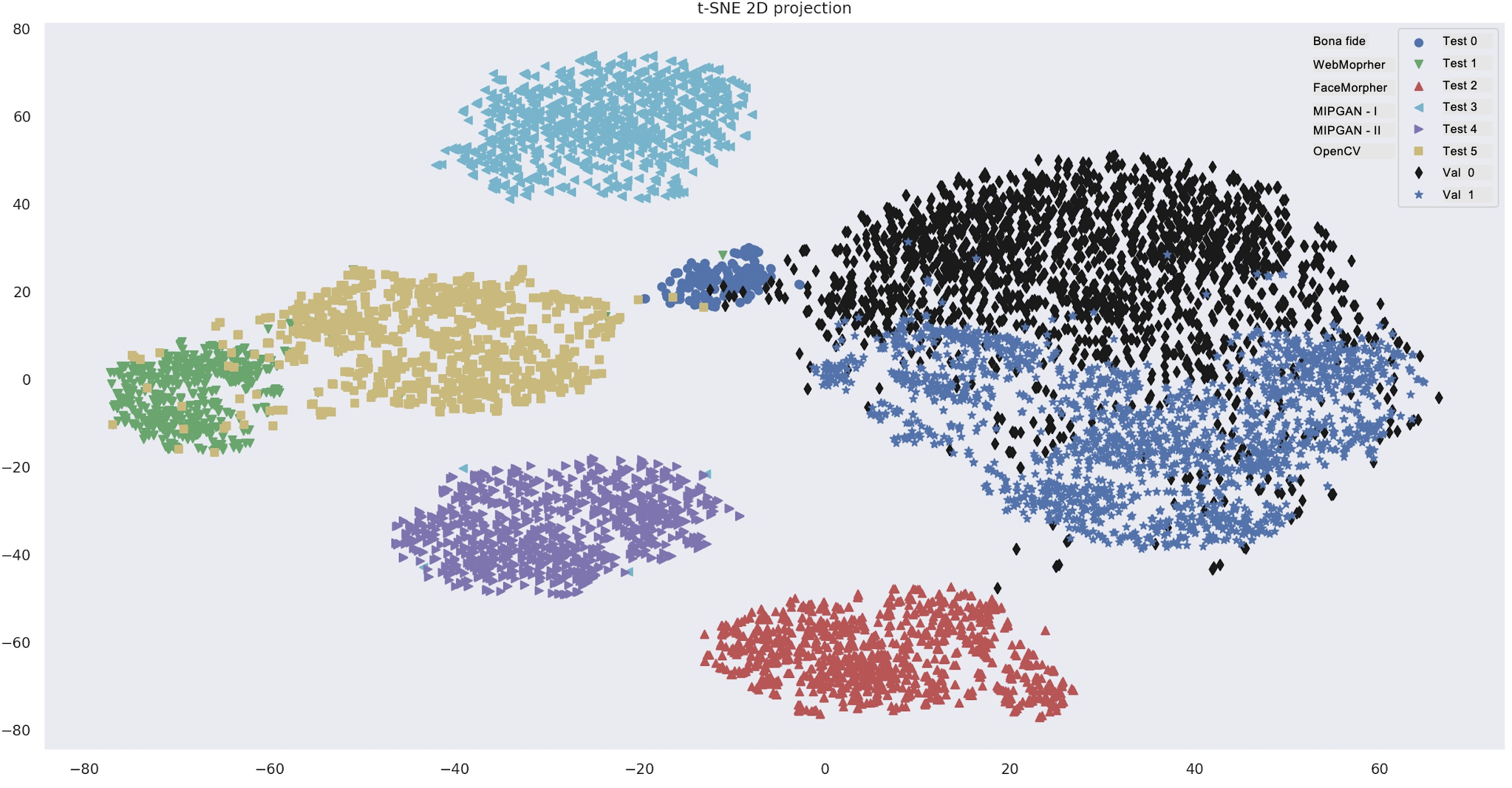}
\caption{\label{tsne-exp1} T-SNE distibution of SDD-test benchmark. Each colour represents a morphing tool. Bona fide (test-0), WebMorpher (test-1), FaceMorpher (test-2), MIPGAN-I (test-3), MIPGAN-II (test-4), OpenCV-Morpher (test-5). Black represents the bona fide validation set (val-0), and Blue (start) represents the morph validation set (val-1).}
\vspace{-1.5em}
\end{figure}

Figure \ref{tsne-exp1} show the T-SNE distribution of the SDD-test Benchmark dataset in a high dimensional space. The blue (star) and black (diamond) colours represent the validation data that belong to the SYN-MAD test dataset. All the images in the validation set are synthetics. The picture also shows that the SDD-test benchmark, even though most of them belong to the class morphs, was depicted in a different high-dimensional space.

\begin{figure}[]
\centering
\includegraphics[scale=0.14]{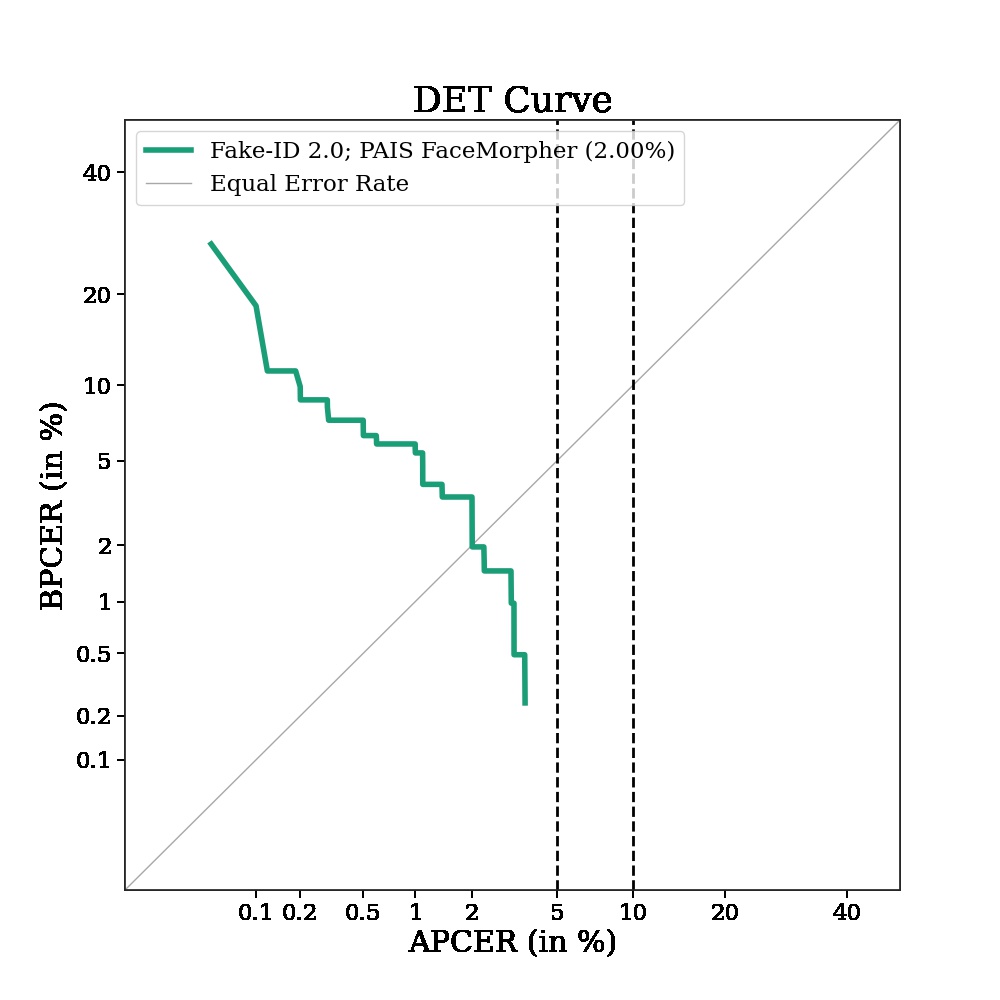}
\includegraphics[scale=0.14]{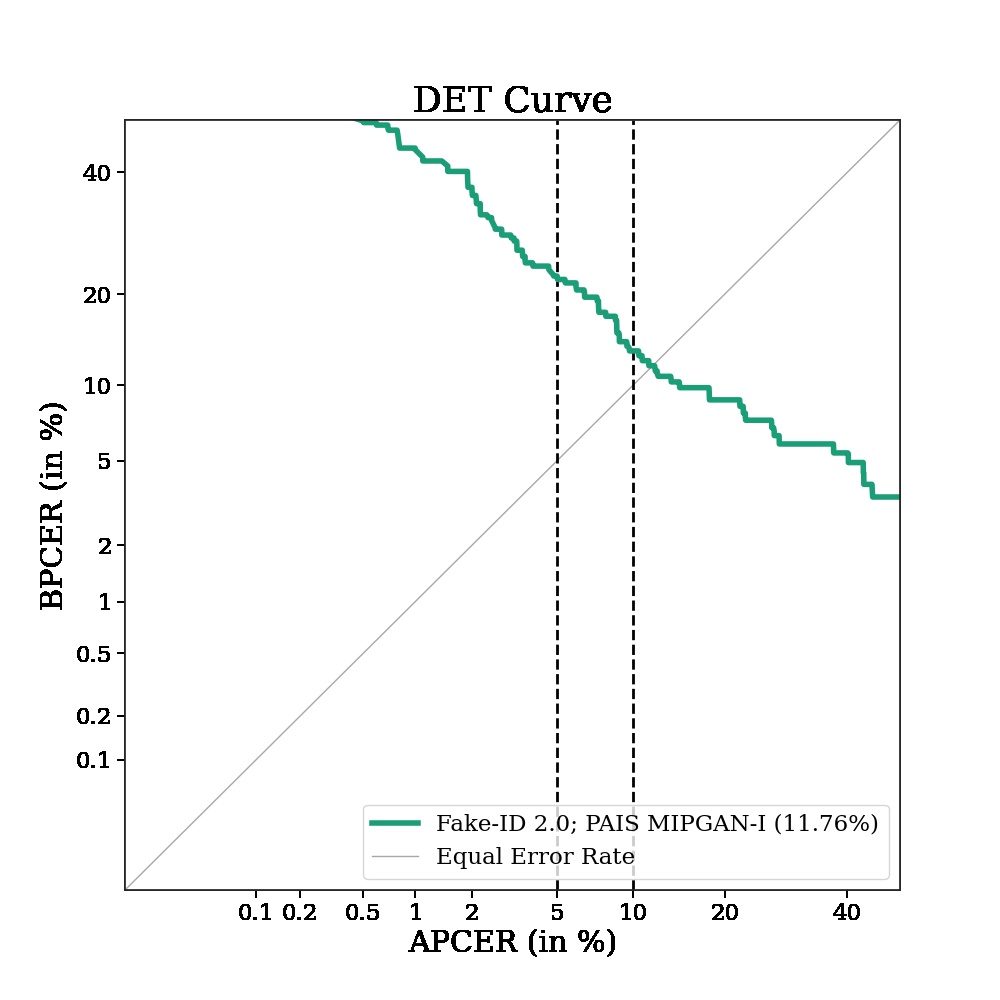}
\includegraphics[scale=0.14]{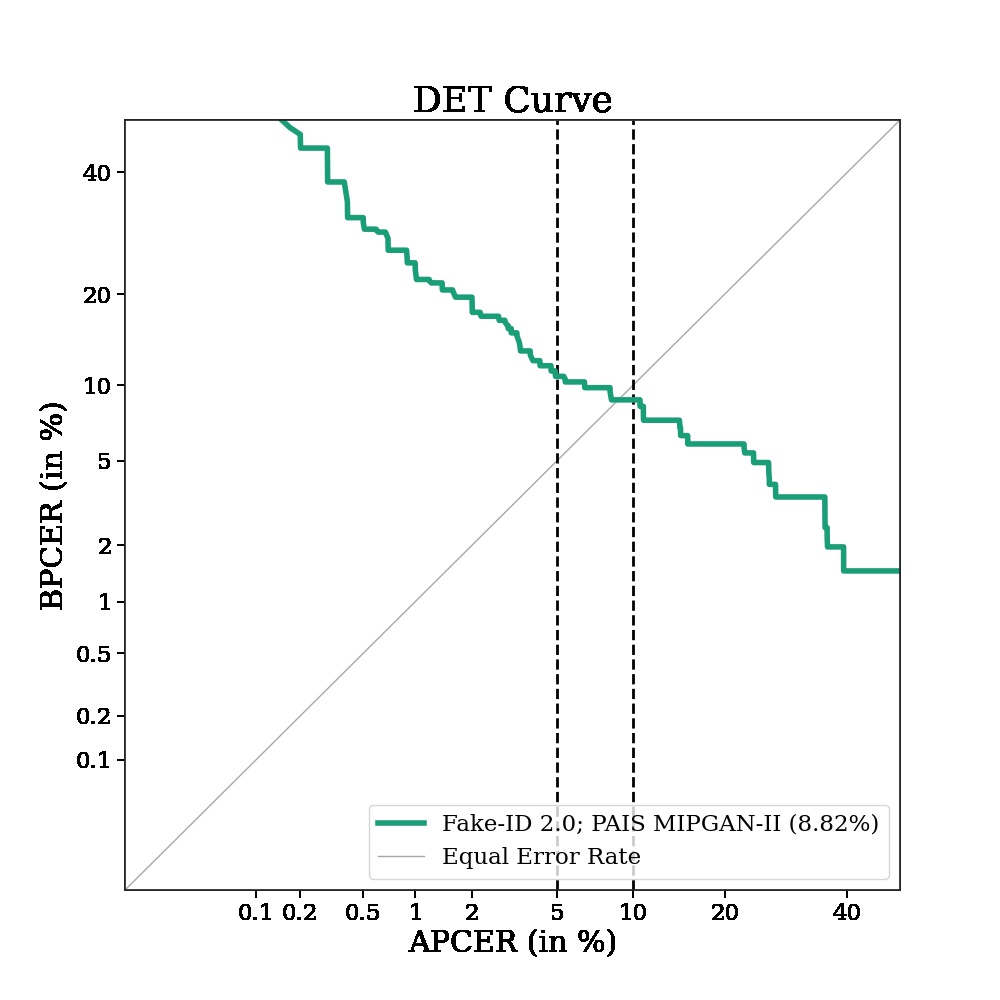} \\
\includegraphics[scale=0.14]{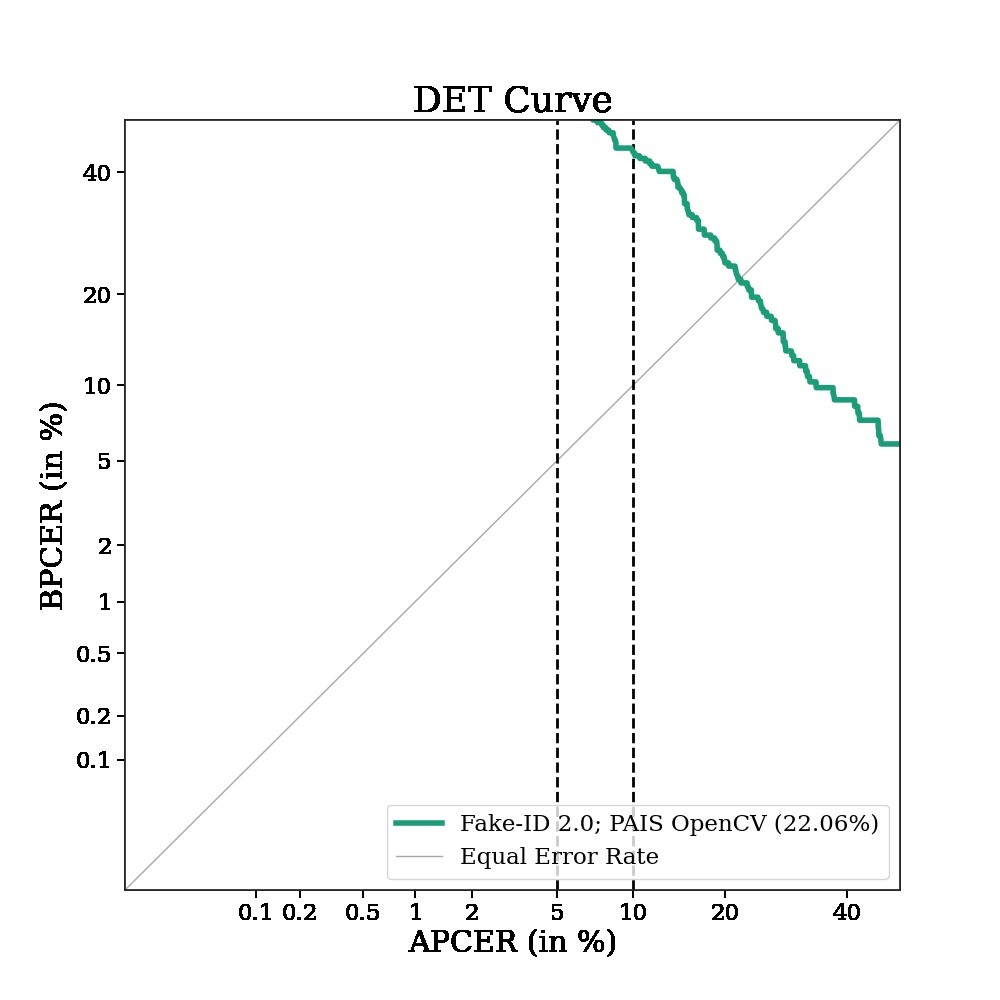}
\includegraphics[scale=0.14]{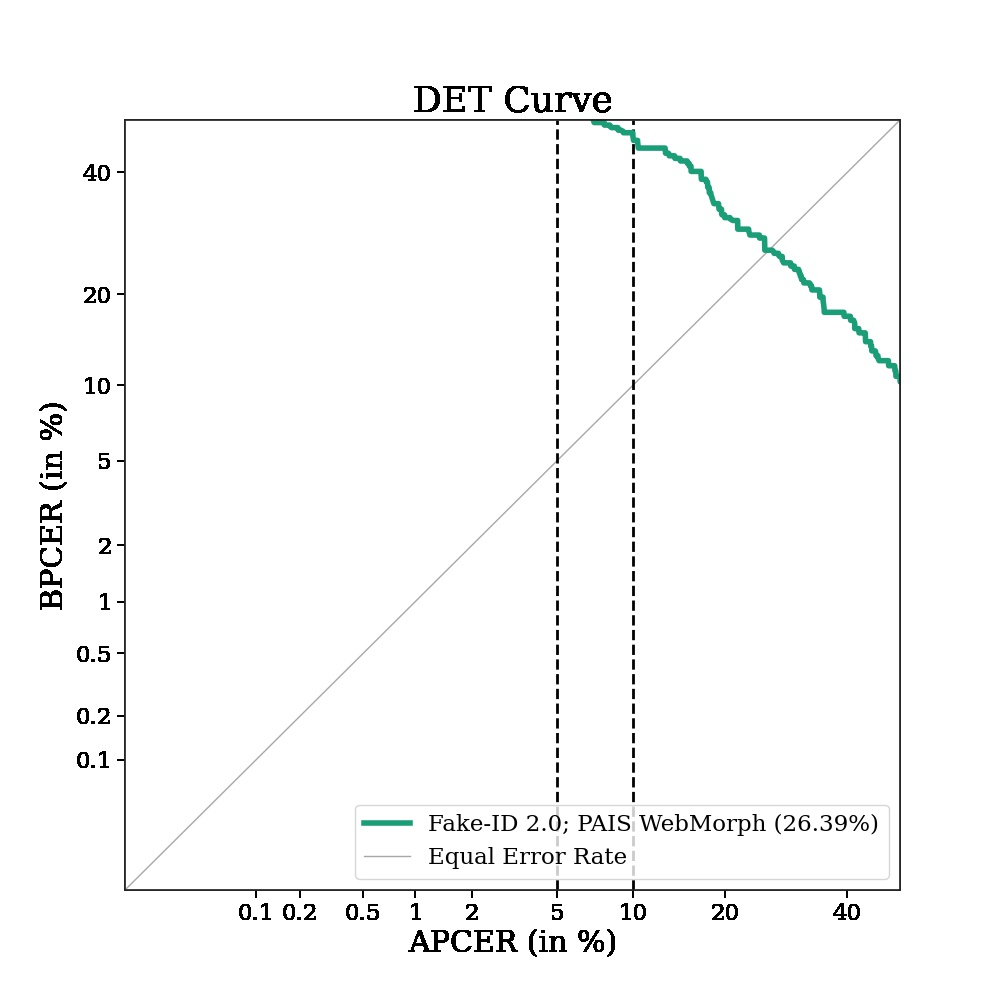}

\caption{\label{det-pais} Exp1: DET curves belong to each class of SDD-Test benchmark. Left to Right: FaceMorpher, MIPGAN-I, MIPGAN-II, OpenCV-Morpher, WebMorpher.  Dot-line indicates BPCER10 and BPCER20, respectively. The EER is reported in parentheses in percentages.}
\vspace{-1.5em}
\end{figure}

Figure \ref{det-pais} show an individual analysis of the SDD benchmark datasets break for each morphing tool separately. Morphed images created by WebMorpher tools have been identified as the most difficult to detect in comparison with FaceMorpher, MIPGAN-I, MIPGAN-II, and OpenCV-Morpher.

\subsection{Exp2 - Train SYN-MAD/TEST SOTA}
The best results from experiment 1 were used for training the EfficientNetB0 again, but now only using the official training set of the SYN-MAD database release for the testing process was evaluated with SOTA: AMSL, FERET, FRGC and FRLL databases. 23,000 bona fide and 13,000 morphed synthetic images were used to train. For the validation set, 4,000 (2,000 bona fide and 2,000 synthetics) images were used.
The best $lr$ was $1e-4$. Table \ref{tab:results_sdd_our} shows a summary with the best results for experiment 2. This experiment shows that synthetic images can not reach generalisation capabilities, and the performance is worse when compared with digital images based on landmark morphing methods. The EER obtained was 29.12\%. Table \ref{tab:results_sdd_our} shows the summary results.

\subsection{Exp3 - Train SOTA/TEST SDD-Benchmark}
For experiment 3, the best results from experiment 1 (EfficientNetB0) also was used for training again, but now used the FERET, FRGC and FRLL images for training and tested in the official test set of the SDD-Benchmark database. The $lr$ was set up to $5e-5$. 
This experiment can infer with a low error the synthetic images even when no synthetic images were used in the training set. The EER, BPCER10 and BPCER20 for SDD-test reached 3.49\%, 1.96\% and 2.94\%, respectively. 
The results for a cross-set with AMSL reached 2.02\% and 3.29\%. Table \ref{tab:results_sdd_our} shows a summary with the best results for experiment 3.

\subsection{Exp4 - Train Mix/TEST SDD-Benchmark}
\begin{figure}[]
\centering
\includegraphics[scale=0.14]{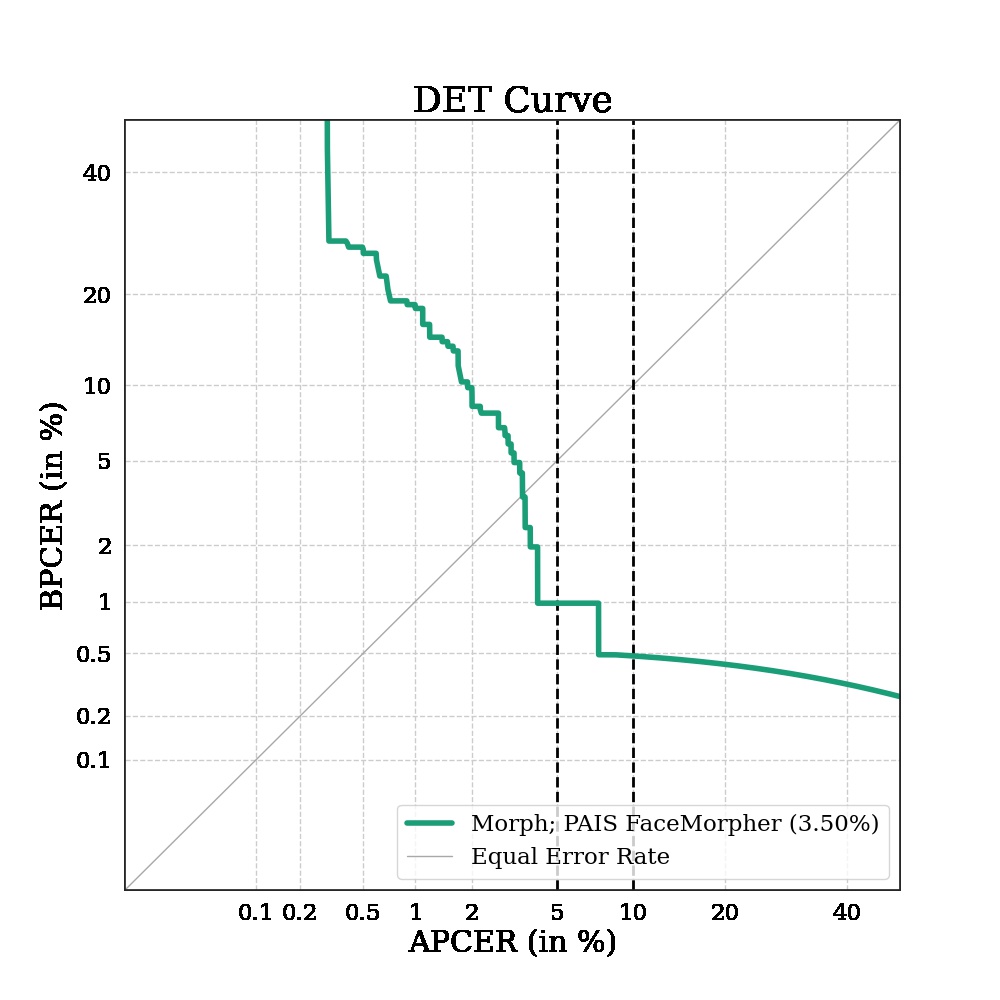}
\includegraphics[scale=0.14]{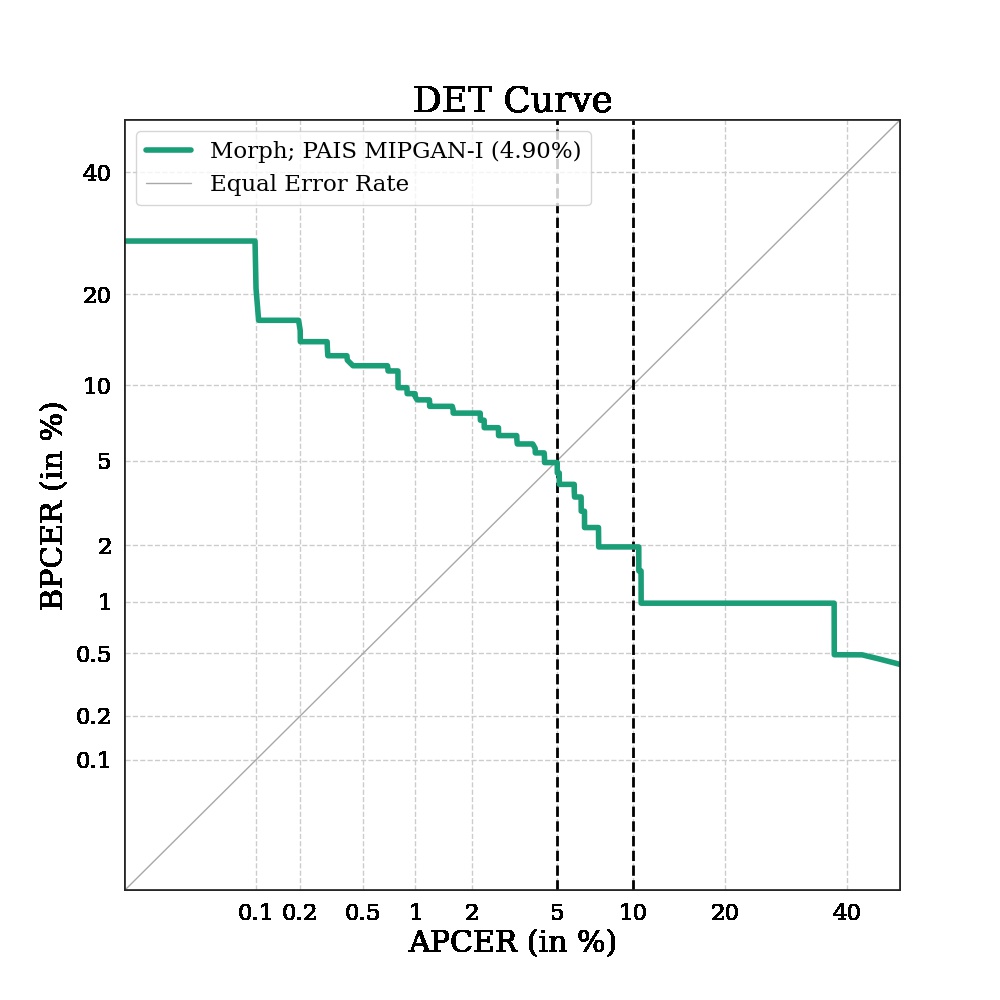}
\includegraphics[scale=0.14] 
{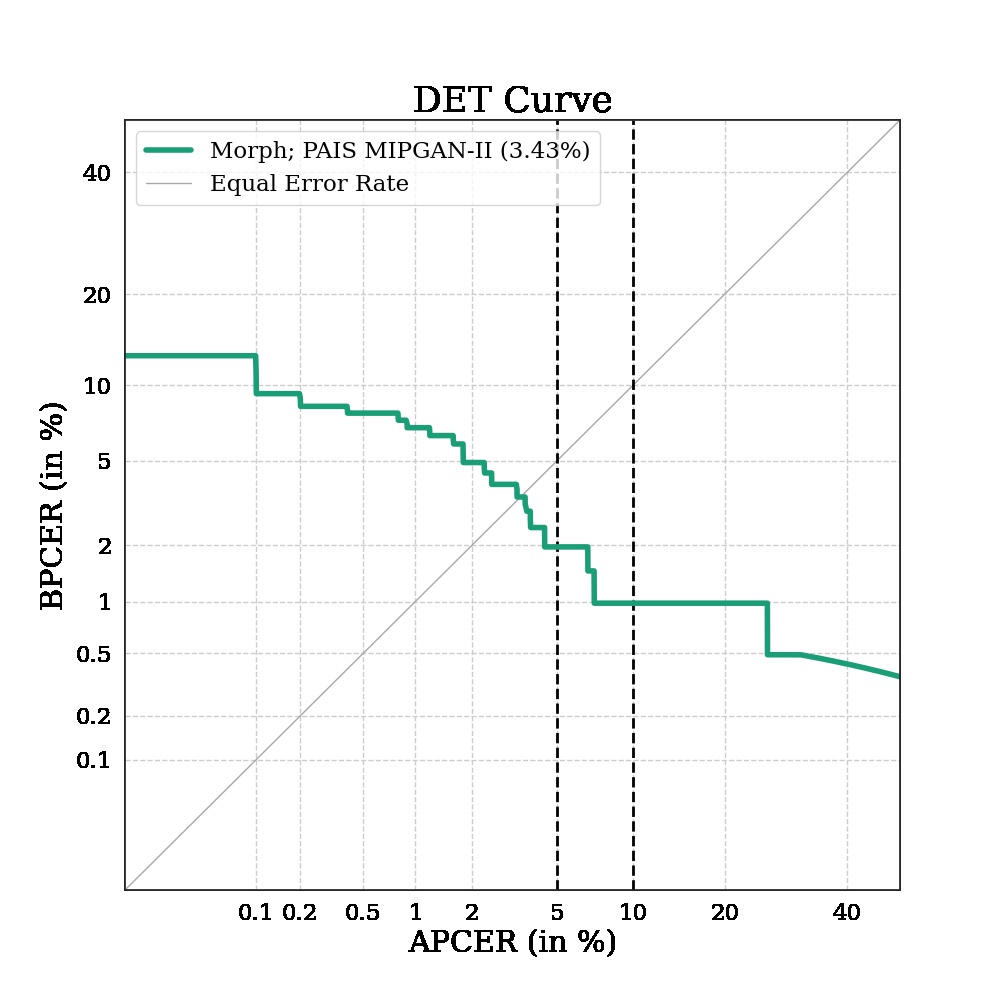} \\
\includegraphics[scale=0.14]{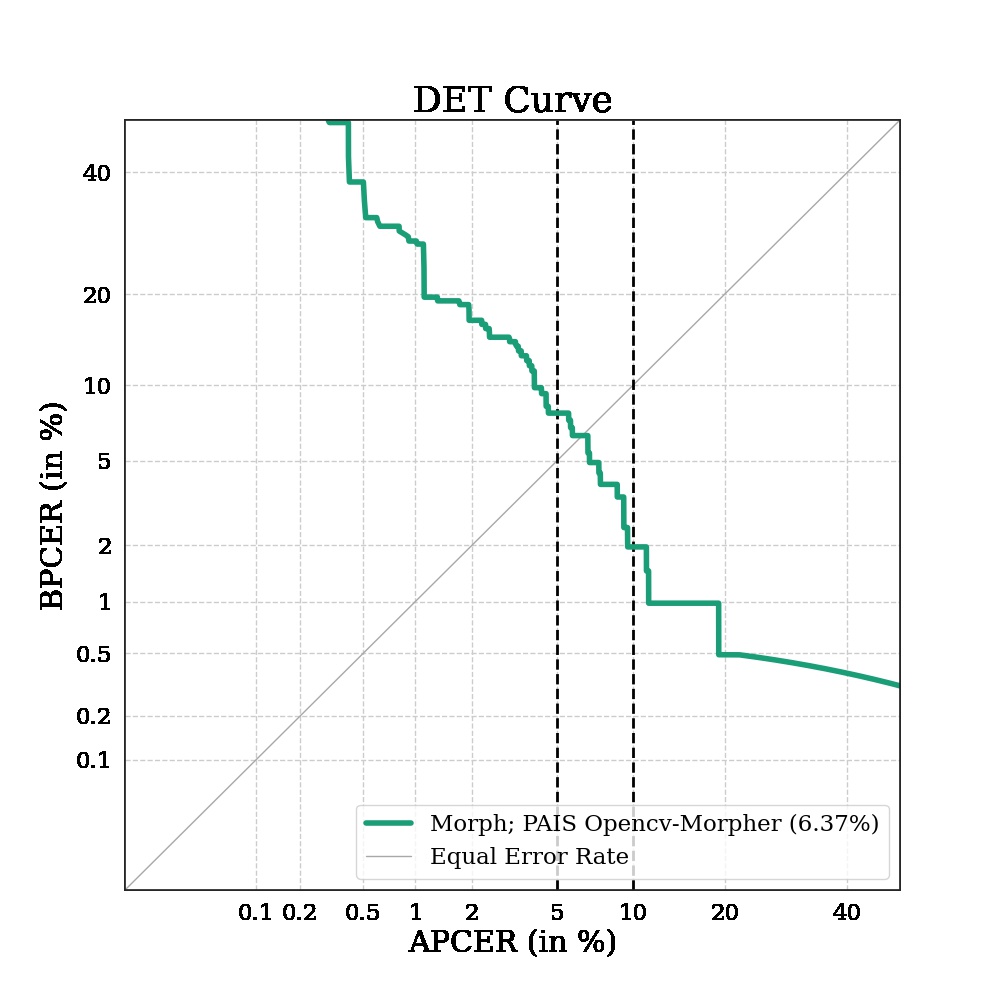}
\includegraphics[scale=0.14]{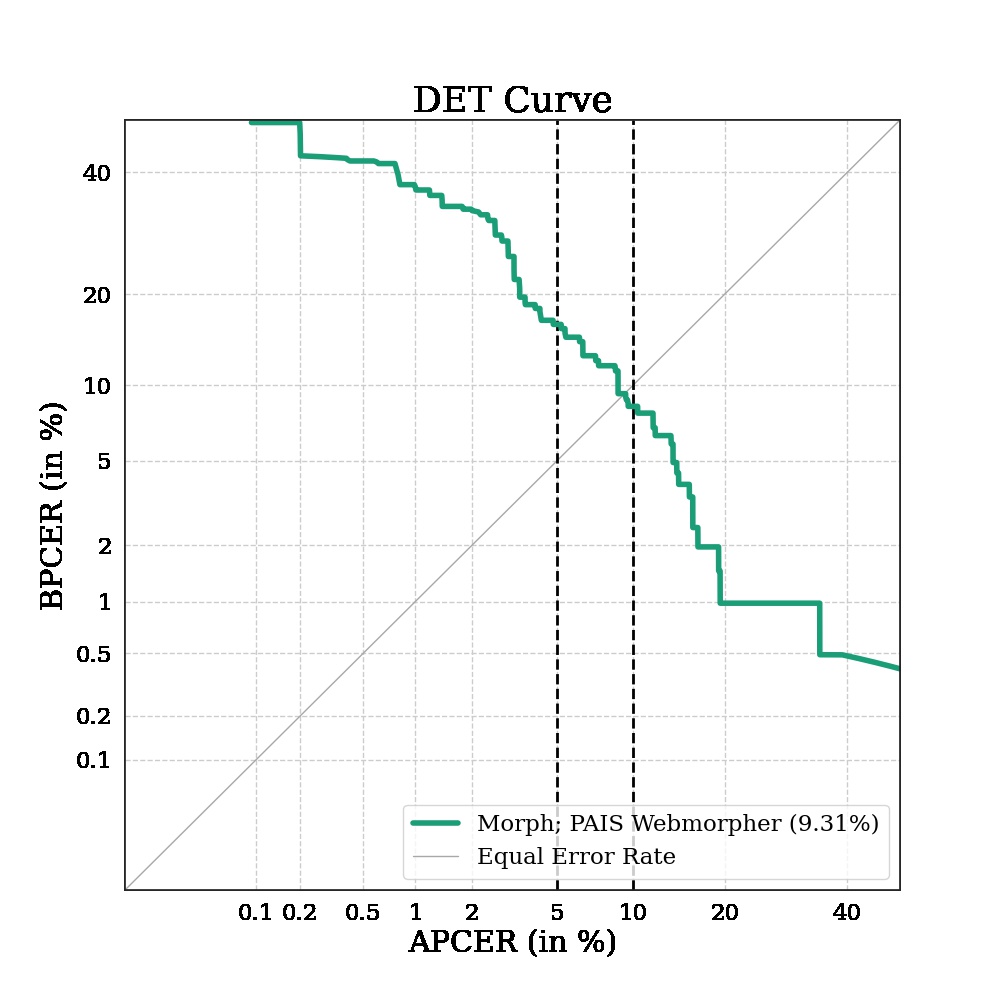}

\caption{\label{det-pais-mix} Exp4: DET curves belong to each class of SDD-Test benchmark, trained on mix-database. Left to Right: FaceMorpher, MIPGAN-I, MIPGAN-II, OpenCV-Morpher, WebMorpher.  Dot-line indicates BPCER10 and BPCER20, respectively. The EER is reported in percentages in parentheses.}
\vspace{-1.5em}
\end{figure}

For Experiment 4, the synthetic images from the SYN-MAD, FERET, and FRGCv2 were mixed by increasing the number of images available for training. The SDD-test benchmark was used as a test set. This experiment allows us to compare our proposed method with the best results of the IJCB 2022 competition obtained by the MorphHRNet team (Table \ref{tab:results_sdd_our}). The best results were obtained when at least 20\% of the images belonged to digital faces (no synthetic). There were 23.000 synthetic face images plus 5,000 FERET and FRGC digital bona fide images and 13,000 morph synthetic images. For the validation set, 2,000 bona fide and 2,000 morphs were used. 
Table \ref{tab:results_sdd_our} shows the summary results. 
Figure \ref{det-pais-mix} shows an individual analysis of the train-mix database evaluated with SDD benchmark datasets break for each morphing tool separately. Morphed images created by WebMorpher tools have been identified as the most difficult to detect in comparison with FaceMorpher, MIPGAN-I, MIPGAN-II, and OpenCV-Morpher. Figure \ref{det-pais-mix} show the improvements reached when the 20\% SOTA database is included in the pure synthetic database. The BPCER10 and BPCER20 for SDD-test reached 0.98\% and 5.39\%, respectively. 

\begin{figure}[]
\centering
\includegraphics[scale=0.22]{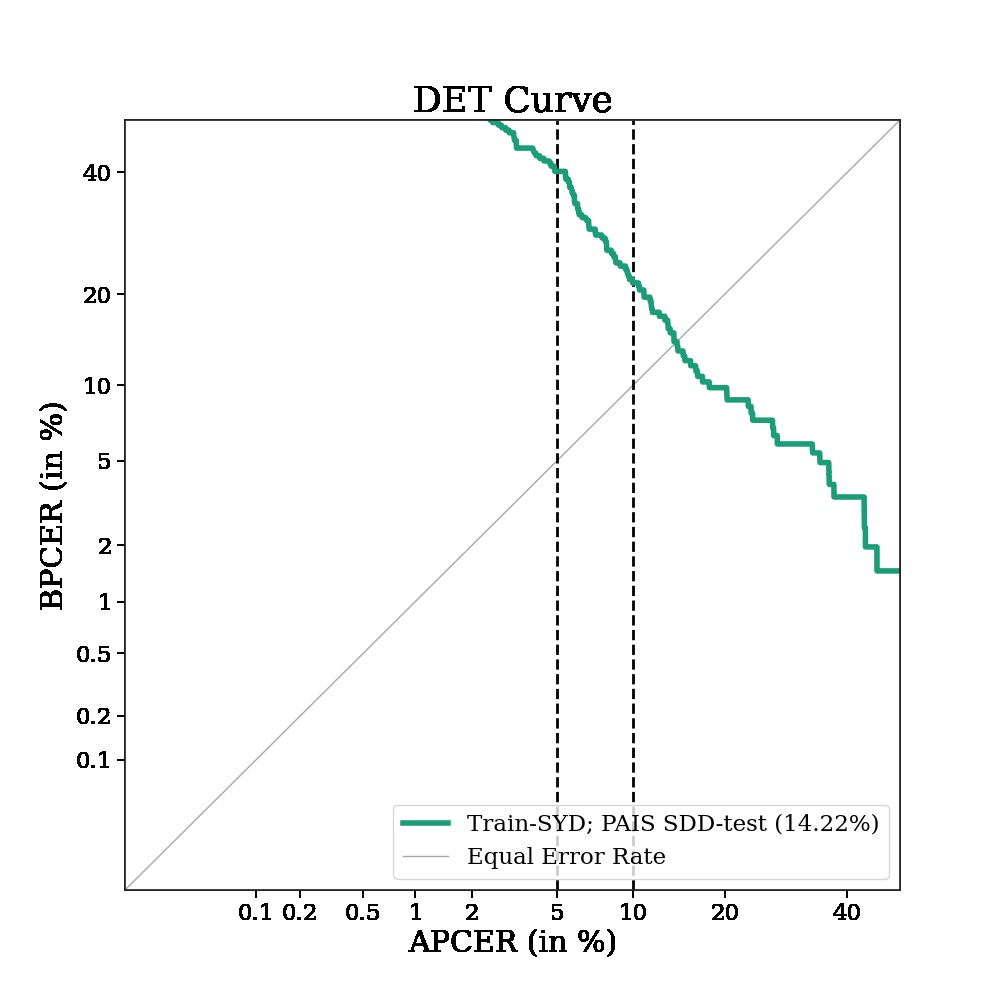}
\includegraphics[scale=0.22]{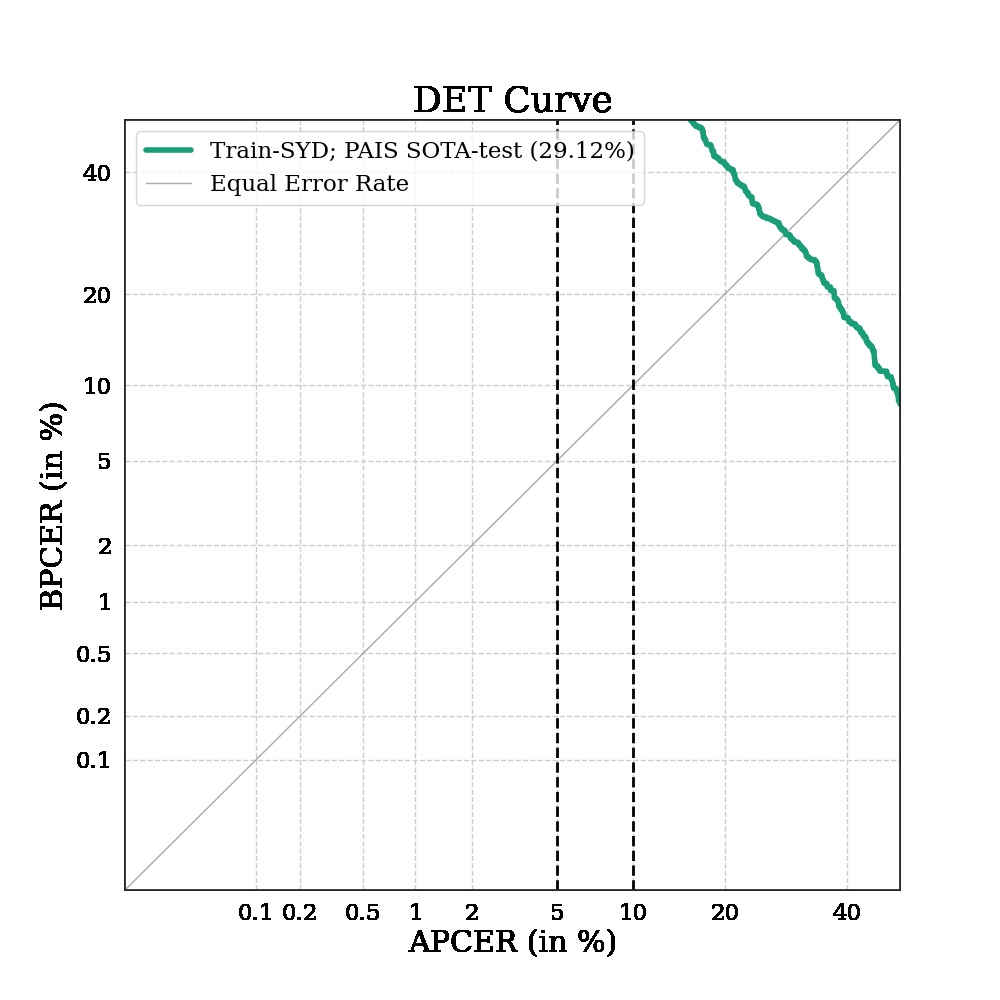}\\
\includegraphics[scale=0.22]{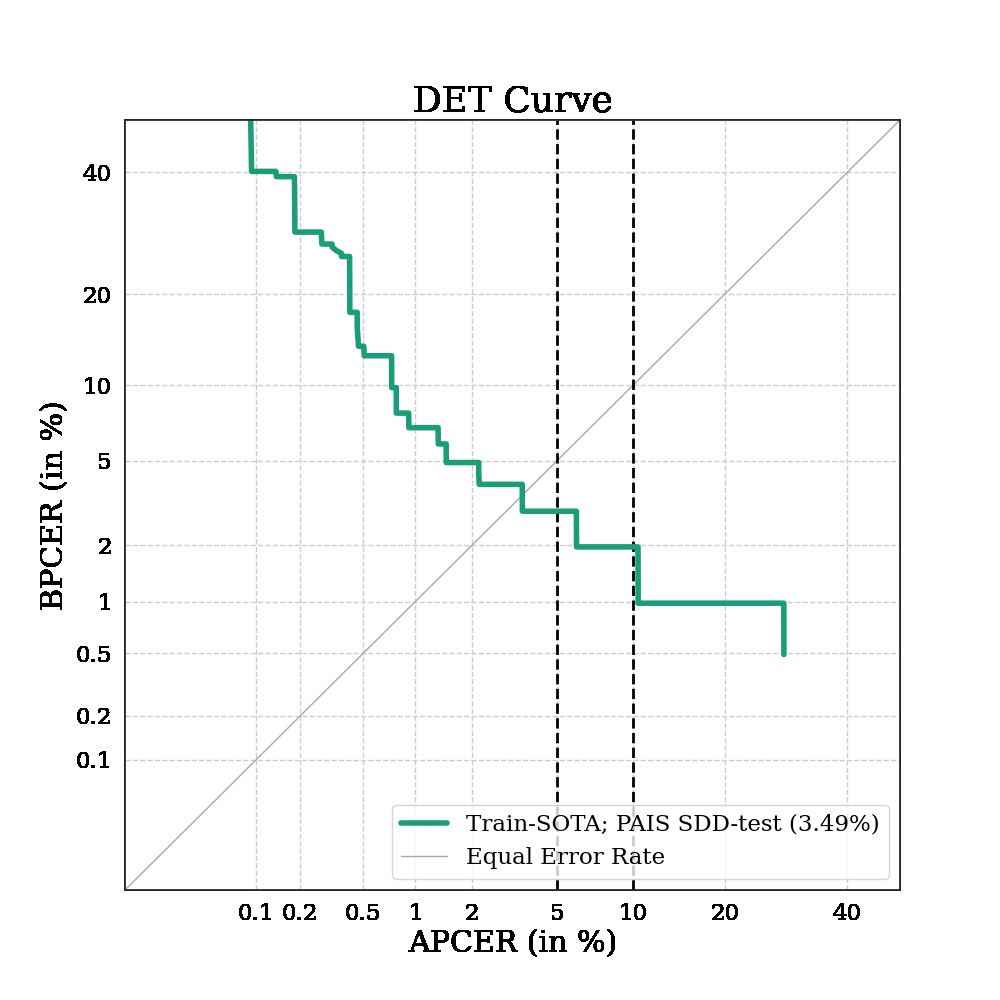}
\includegraphics[scale=0.22]{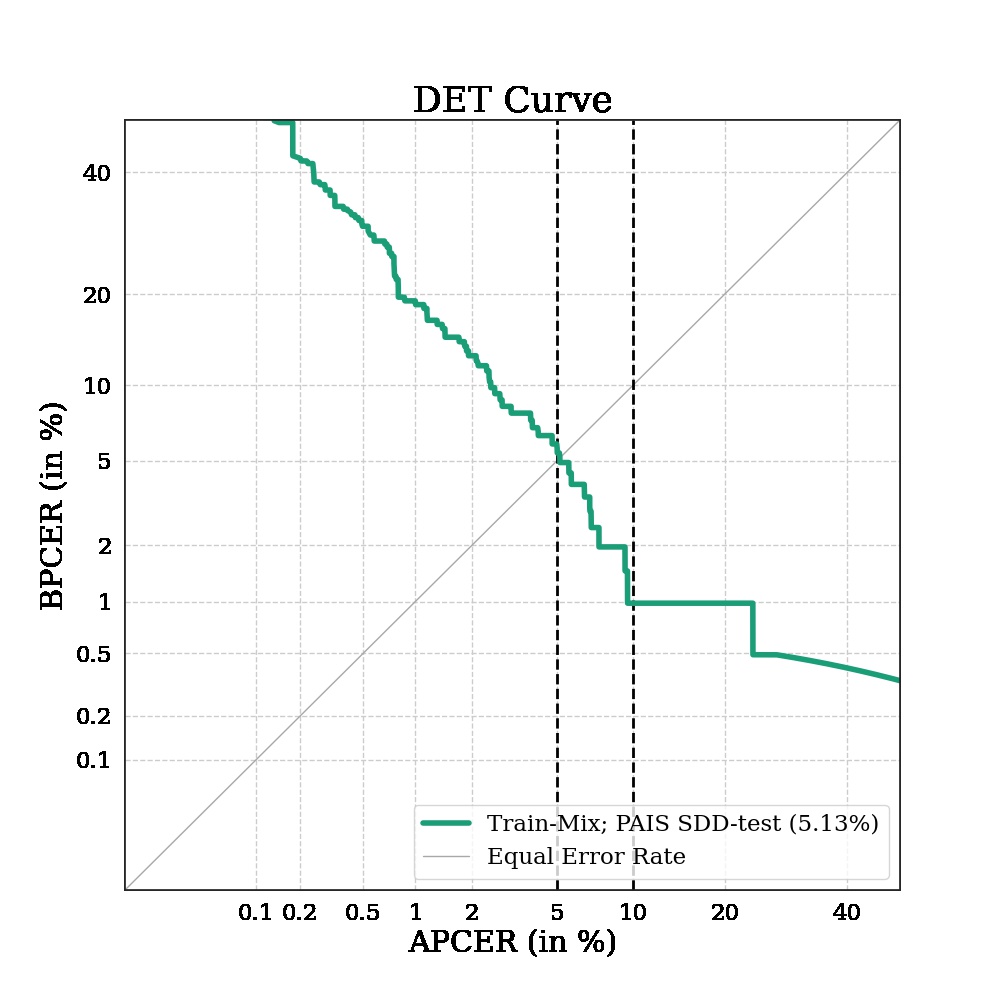}

\caption{\label{test3_v1} DET curves. Left to Right: Exp1: Trained with SYN-MAD/test SDD-Benchmark. Exp2: Trained with SYN-MAD/test-SOTA. Exp3: Trained with SOTA/test SDD-Benchmark. Exp4: Trained with Mix database-test SDD-Benchmark. Dot-line indicates BPCER10 and BPCER20, respectively. The EER is reported in percentages in parentheses.}
\vspace{-1.5em}
\end{figure}

Figure \ref{test3_v1} shows DET curves with the results when the test datasets are fully evaluated. This means all the morphed images belong to the benchmark dataset together as one morph class. We can observe that the performance of Exp-1, the SDD-test benchmark reached a higher EER of 14.22\%. This result is valuable because it was obtained purely with synthetic images used as morph and bona fide. Also, the results show that controlling the quality of the synthetic images used for this process is necessary before creating the morph.  Some images from SYN-MAD present relevant artefacts in some face areas. Then, these artefacts are also translated into morphed images. Exp-2 reached an EER of 3.49\% which represents a third party when using only synthetic images. For Exp-4, which is based on a mixed dataset, the EER obtained was 5.13\%.
\vspace{-0.3cm}

\section{Conclusions}
\label{sec:conclusions}

This work assesses the impact of using purely synthetic images to train MAD. Our proposal for a detection approach is based on a Siamese network with a semi-hard-loss function and outperforms the best results of the IJCB2021 competition. The results show that performance based only on synthetic images, even reaching competitive results, cannot reach the best results and totally discards digital images. The main lack regarding the SOTA database available was only an increased number of digital morphing tools despite bona fide and the fact that print/scan scenarios are one of the most realistic in border gate operation. However, developing this kind of database is time-consuming. 

\subsubsection{Acknowledgements} This work is supported by the European Union’s Horizon 2020 research and innovation program under grant agreement No 883356 and the German Federal Ministry of Education and Research and the Hessen State Ministry for Higher Education, Research and the Arts within their joint support of the National Research Center for Applied Cybersecurity ATHENE.

\bibliographystyle{splncs04}
\bibliography{main}

\end{document}